\documentclass{article}

\usepackage{microtype}
\usepackage{graphicx}
\usepackage{subfigure}
\usepackage{booktabs} 
\usepackage{url}

\usepackage{hyperref}

\usepackage[accepted]{icml2023}

\usepackage{amsmath}
\usepackage{amssymb}
\usepackage{mathtools}
\usepackage{amsthm}

\usepackage[capitalize,noabbrev]{cleveref}

\theoremstyle{plain}
\newtheorem{theorem}{Theorem}[section]
\newtheorem{proposition}[theorem]{Proposition}

\theoremstyle{definition}

\theoremstyle{remark}

\usepackage[textsize=tiny]{todonotes}

\usepackage{amsmath,amsfonts,bm}
\usepackage{multirow}
\usepackage{tabularx}
\usepackage{graphicx}
\usepackage{adjustbox}
\usepackage{xspace}

\newcommand{\dif}{\Phi}
\newcommand{\ftd}{{\scriptscriptstyle \text{FTD}}}

\usepackage[algo2e,ruled,vlined]{algorithm2e}

\newcommand{\mset}[1]{\left\{\kern-.5em\left\{ #1 \right\}\kern-.5em\right\}}
\newcommand{\mmset}[1]{\{\kern-.4em\{ #1 \}\kern-.4em\}}

\newcommand{\norm}[1]{\left\Vert#1\right\Vert}

\newcommand{\set}[1]{\left\{#1\right\}}

\newcommand{\Real}{\mathbb R}

\newcommand{\too}{\rightarrow}

\newcommand{\eg}{{e.g.}\xspace}
\newcommand{\ie}{{i.e.}\xspace}

\def\eqref#1{Eq.~\ref{#1}}

\def\1{\bm{1}}

\def\vec1{{\bm{1}}}

\DeclareMathAlphabet{\mathsfit}{\encodingdefault}{\sfdefault}{m}{sl}
\SetMathAlphabet{\mathsfit}{bold}{\encodingdefault}{\sfdefault}{bx}{n}

\def\gC{{\mathcal{C}}}

\def\gI{{\mathcal{I}}}
\def\gJ{{\mathcal{J}}}

\def\gL{{\mathcal{L}}}

\def\gY{{\mathcal{Y}}}
\def\gZ{{\mathcal{Z}}}

\DeclareMathOperator*{\argmin}{arg\,min}

\usepackage{arydshln}
\usepackage{wrapfig,lipsum,booktabs}
\usepackage{sidecap}
\sidecaptionvpos{figure}{t}

\usepackage{lipsum}
\newcommand\blfootnote[1]{%
  \begingroup
  \renewcommand\thefootnote{}\footnote{#1}%
  \addtocounter{footnote}{-1}%
  \endgroup
}

\begin{document}

\twocolumn[
\icmltitle{MultiDiffusion:  Fusing Diffusion Paths for Controlled Image Generation}



\icmlsetsymbol{equal}{*}

\begin{icmlauthorlist}
\icmlauthor{Omer Bar-Tal}{equal}
\icmlauthor{Lior Yariv}{equal}
\icmlauthor{Yaron Lipman}{}
\icmlauthor{Tali Dekel}{}
\end{icmlauthorlist}



\icmlkeywords{Machine Learning, ICML}

\vskip 0.3in
]



\printAffiliationsAndNotice{\icmlEqualContribution} 

\begin{figure*}[ht!]
    \centering
    \includegraphics[width=1.\textwidth]{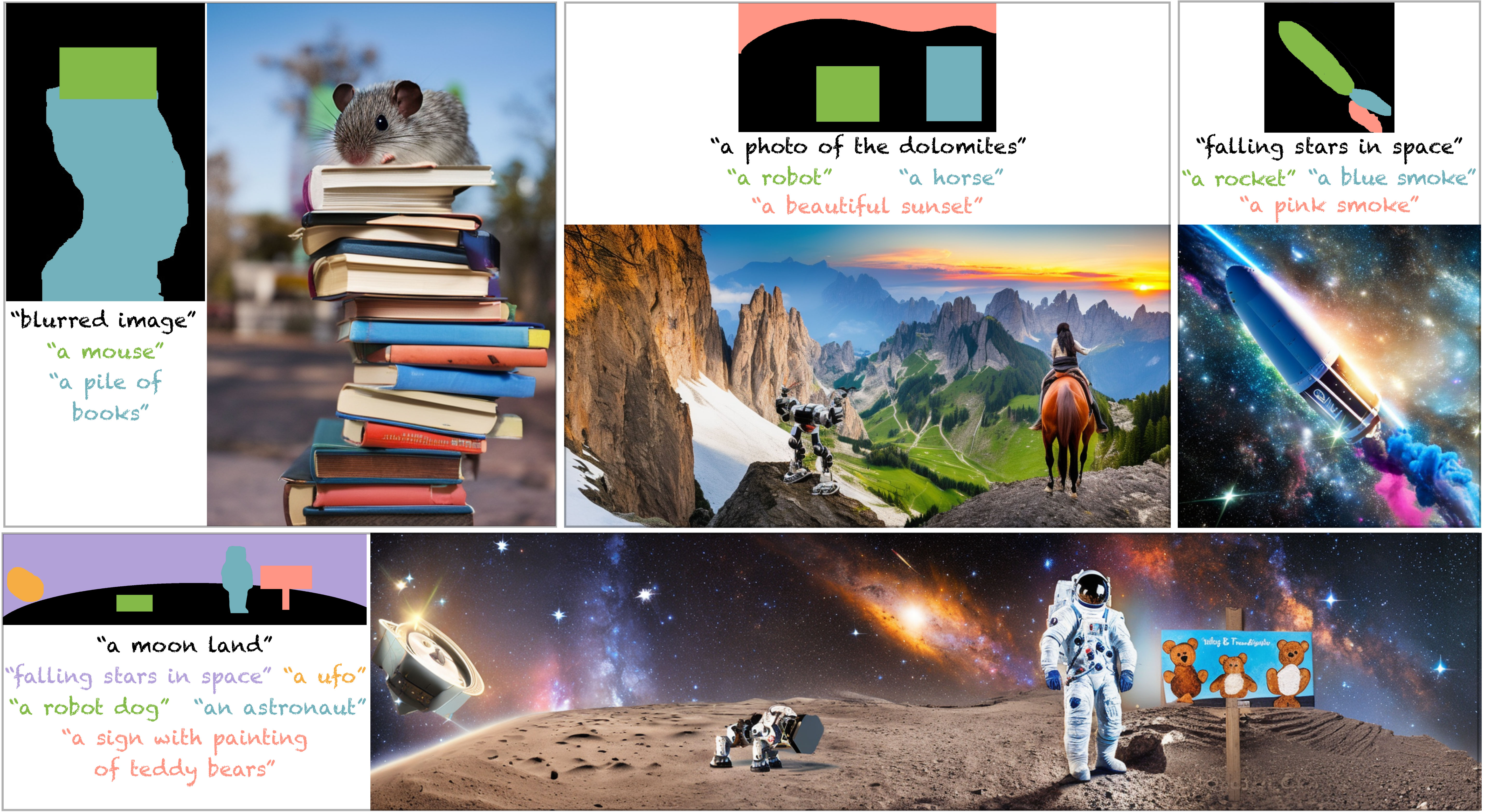} \hspace{-10pt}
    \caption{MultiDiffusion enables flexible  text-to-image generation, unifying multiple controls over the generated content, including desired aspect ratio, or simple spatial guiding signals such as rough region-based text-prompts.}
    \label{fig:teaser}
\end{figure*}

\begin{abstract}
Recent advances in text-to-image generation with diffusion models present transformative capabilities in image quality. However, user controllability of the generated image, and fast adaptation to new tasks still remains an open challenge, currently mostly addressed by costly and long re-training and fine-tuning or ad-hoc adaptations to specific image generation tasks. In this work, we present \emph{MultiDiffusion}, a unified framework that enables versatile and controllable image generation, using a pre-trained text-to-image diffusion model, without any further training or finetuning.  At the center of our approach is a new generation process, based on an optimization task that binds together multiple diffusion generation processes with a shared  set of parameters or constraints. 
We show that  MultiDiffusion can be readily applied to generate high quality and diverse images that adhere to user-provided controls, such as desired aspect ratio (e.g., panorama), and spatial guiding signals, ranging from tight segmentation masks to bounding boxes. 
\end{abstract}

\section{Introduction}\label{sec:intro}
Text-to-image generative models have emerged as a ``disruptive technology'',  demonstrating unprecedented capabilities in synthesizing high-quality and diverse images from text prompts, where diffusion models are currently established as state-of-the-art \cite{saharia202_imagen,ramesh2022dalle2,rombach2022high,croitoru2022diffusion}.  While this progress holds a great promise in changing the way we can create digital content, deploying text-to-image models to real-world applications remains challenging due to the difficulty to provide users with intuitive control over the generated content. Currently, controllability over diffusion models is achieved in one of two ways: (i) training a model from scratch or finetuning a given diffusion model for the task at hand (\eg, inpainting, layout-to-image training, etc.  \cite{wang2022piti,ramesh2022dalle2,rombach2022high,nichol2021glide,avrahami2022spatext,brooks2022instructpix2pix,wang2022semantic}). With the ever-increasing scale of models and training data, this approach often requires \emph{extensive compute} and \emph{long development period}, even in a finetuning setting.
(ii) Reuse a pre-trained model and add some controlled generation capability. Previously, these methods have concentrated on specific tasks and designed a tailored methodology (\eg, replacing objects in an image, manipulating style, or controlling layout \cite{pnpDiffusion2022,hertz2022prompt_to_prompt,avrahami2022blended}). 
\blfootnote{Project page is available at \url{https://multidiffusion.github.io}.}

The goal of this work is to design \emph{MultiDiffusion}, a new unified framework that significantly increases the flexibility in adapting a pre-trained (reference) diffusion model to controlled image generation. The basic idea behind the MultiDiffusion is to define a \emph{new generation process} that is composed of several reference diffusion generation processes binded together with a set of shared parameters or constraints. In more detail, the reference diffusion model is applied to different regions in the generated image, predicting a denoising sampling step for each. In turn, the MultiDiffusion takes a global denoising sampling step reconciling all these different steps via least squares optimal solution. 

\begin{figure*}[ht!]
    \centering
    \includegraphics[width=1\textwidth]{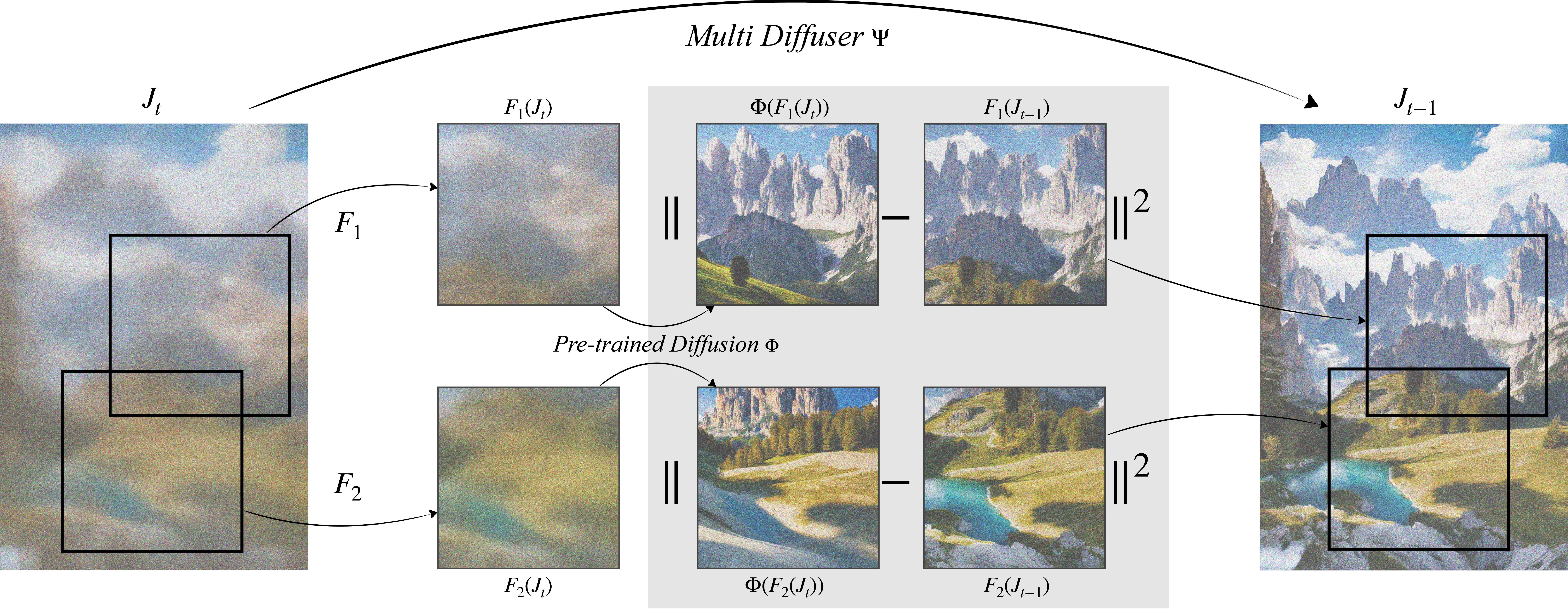}\hspace{-10pt}
    \caption{MultiDiffusion: a new generation process, $\Psi$, is defined over a pre-trained reference model $\dif$. Starting from a noise image $J_T$, at each generation step, we solve an optimization task whose objective is that \emph{each} crop $F_i(J_t)$ will follow as closely as possible its denoised version $\dif(F_i(J_t))$. Note that while each denoising step $\dif(F_i(J_t))$ may pull to a different direction, our process fuses these inconsistent directions into a \emph{global} denoising step $\Phi(J_t)$, resulting in a high-quality seamless image.}
    \label{fig:pipeline}
\end{figure*}

For example, consider the task of generating an image at arbitrary aspect ratio given a reference diffusion model trained on square images (Fig.~\ref{fig:pipeline}). At each denoising step, the MultiDiffusion fuses the denoising directions, provided by the reference model, from \emph{all} the square crops, and strives to follow them all as closely as possible, constrained by the fact that nearby crops share common pixels. Intuitively, we encourage each crop to be a real sample from the reference model. Note that while each crop might pull to a different denoising direction, our framework yields a unified denoising step, hence produces high-quality and seamless images.

With MultiDiffusion, we are able to harness a reference pre-trained text-to-image model to different applications including synthesizing images at desired resolution or aspect ratio, or synthesizing images using rough region-based text prompts, as seen in Fig.~\ref{fig:teaser}.  Notably, our framework allows to solve these tasks \emph{simultaneously},  using a common generation process.  Comparing to relevant baselines, we found that our approach is able to produce state-of-the-art controlled generation quality even compared to methods that are specifically trained for these tasks. Furthermore, our method works efficiently,  without introducing computational overhead.  \vspace{-5pt}

\section{Related Work}
\paragraph{Diffusion Models}
\label{sec:related_diffusion} Diffusion models \cite{sohl2015deep,croitoru2022diffusion,beatgan,ddpm,nichol2021improved} are a class of generative probabilistic models that aim to approximate a data distribution $q$, and are easy to sample from. Specifically, these models take a Gaussian noise input $I_T\sim \mathcal{N}(0,I)$, and through a series of gradual denoising steps, transform it into a sample $I_0$, that should be distributed according to $q$. The number of denoising steps, and the parameterization of the transformation varies among different works \cite{sohl2015deep,ddpm,song2020denoising,lu2022dpm,lu2022dpm_2,liu2022pseudo}.
Recently, Diffusion Models have emerged as state-of-the-art generators due to their success in learning complex distributions and generating diverse high quality samples. These models have been successfully used in various domains, including images \cite{ beatgan,nichol2021improved,saharia202_imagen,ramesh2022dalle2,rombach2022high}, video \cite{ho2022imagen_video,make_a_video}, 3D scenes \cite{mueller2022diffrf}, and motion sequences \cite{yuan2022physdiff,tevet2022human}.

\paragraph{Controllable generation with diffusion models}
\label{sec:related_control}
Diffusion models can be trained with guiding input channels (\eg, semantic layout, category label) and successfully perform conditional image generation \cite{ramesh2021dalle1,saharia2022_diffusion_sr,saharia2022palette,wang2022piti,preechakul2022diffusion,ho2022classifier}. The most prominent example of conditional diffusion models is recent text-to-image diffusion models, which have demonstrated groundbreaking synthesis capabilities \cite{nichol2021glide,saharia202_imagen,ramesh2022dalle2,nichol2021glide,rombach2022high,sheynin2022knn}. However, these models provide only little control over the generated content, which is mainly achieved through the input text. Recently, a surge of methods have been proposed to gain wider and better user controllability.
Existing methods can be roughly divided into two  main approaches: 
(i) methods that incorporate explicit control by using additional guiding signals to the model \cite{avrahami2022spatext,rombach2022high,brooks2022instructpix2pix}. 
However, these works require costly extensive training on curated datasets. (ii) On the other side of the spectrum, numerous methods proposed to implicitly control the generated content by manipulating the generation process of a pre-trained model \cite{kwon2022diffusion_splice,meng2021sdedit,pnpDiffusion2022,hertz2022prompt_to_prompt,avrahami2022blended_1,choi2021ilvr,mokady2022null,couairon2022diffedit,kong2023leveraging,kwon2022diffusion}
or by performing lightweight model finetuning \cite{ruiz2022dreambooth,kawar2022imagic,kim2022diffusionclip,valevski2022unitune}. Avarahami \textit{et al.} designed image inpainting methods \cite{avrahami2022blended,avrahami2022blended_1} that do not require finetuning. 
Recent works \cite{pnpDiffusion2022,hertz2022prompt_to_prompt} rely on architectural properties and insights about the internal features of the pretrained model, and tailor image editing techniques accordingly. 
Our work also manipulates the generation process of a pretrained diffusion model, and does not require any training or finetuning. However, in contrast to existing works that target a specific application, without a well defined objective, we propose a more general approach that allows us to unify different user control inputs in a more principled manner.

\section{Method}
\label{sec:method}
We consider a pre-trained diffusion model, which serves as a reference model: $$\dif:\gI\times \gY \too \gI$$ working in image space $\gI=\Real^{H\times W \times C}$ and condition space $\gY$, \eg, $y\in \gY$ is a text prompt. Initializing $I_T\sim P_\gI$, where $P_\gI$ represents the distribution of Gaussian i.i.d. pixel values, and setting a condition $y\in \gY$, the diffusion model builds a sequence of images, 
\begin{equation}\label{e:Phi_series}
\begin{array}{lcr}
     I_T, I_{T-1}, \ldots ,I_0 & \text{s.t.}& I_{t-1}=\dif(I_t|y) 
\end{array}
\end{equation}
gradually transforming the noisy image $I_T$ into a clean image $I_0$.

\paragraph{MultiDiffusion.} Our goal is to leverage $\dif$ to generate images in a potentially different image space $\gJ=\Real^{H'\times W' \times C}$ and condition space $\gZ$,  without any training or finetuning. To do so, we define a \emph{MultiDiffusion process}, defined by a function, called \emph{MultiDiffuser}, $$\Psi:\gJ\times \gZ\too \gJ$$
The MultiDiffusion, similarly to a diffusion process, starts with some initial noisy input $J_T\sim P_\gJ$, where $P_\gJ$ is a noise distribution over $\gJ$, and produces a series of images
\begin{equation}\label{e:Psi_series}
\begin{array}{lcr}
     J_T, J_{T-1}, \ldots ,J_0 & \text{s.t.}& J_{t-1}=\Psi(J_t|z) 
\end{array}
\end{equation}

Our key idea is to define $\Psi$ to be \emph{as-consistent-as-possible} with $\dif$. More specifically, we define a set of mappings between the target and reference image spaces $F_i:\gJ\too \gI$, and a corresponding set of mappings between the condition spaces: $\lambda_i:\gZ\too \gY$ where $i\in [n]=\set{1,\ldots,n}$. These mappings are application depended, as will be described later in Sec.~\ref{sec:applications}. Our goal is  to make every MultiDiffuser step $J_{t-1}=\Psi(J_t|z)$ follow as closely as possible $\dif(I^i_t | y_i)$, $i\in[n]$, i.e., the denoising steps of $\dif$ when applied to the images and conditions:
\begin{equation*}
    I^i_t=F_i(J_t), \quad y_i=\lambda_i(z)
\end{equation*}
Formally, our new process is given by solving the following optimization problem:
\definecolor{mygray}{gray}{0.95}
\begin{center}\vspace{-10pt}			
    \colorbox{mygray} {		
      \begin{minipage}{0.977\linewidth} 	
       \centering
        \begin{equation}\label{e:Psi}
    \begin{aligned}
    \Psi(J_t|z)  = &\argmin_{J\in \gJ}\ \  \gL_{\ftd}(J|J_t,z) \\
    \end{aligned} 
\end{equation}
      \end{minipage}}			
\end{center}

\begin{equation}\label{e:ftd}
    \gL_{\ftd}(J | J_t,z) = \sum_{i=1}^n \Big \| W_i \otimes \Big [ F_{i}(J) - \Phi(I^i_t|y_i) \Big ] \Big \|^2
\end{equation}
where $W_i\in \Real^{H\times W}_+$ are per pixel weights and $\otimes$ is the Hadamard product. 
Intuitively, the FTD loss reconciles, in the least-squares sense, the different denoising sampling steps, $\Phi(I^i_t|y_i)$, suggested on different regions, $F_i(J_t)$, of the generated image $J_t$. Fig.~\ref{fig:pipeline} illustrates one step of the MultiDiffuser; Algorithm \ref{alg:FDS} recaps the MultiDiffusion sampling process.

\paragraph{Closed-form formula.} In the applications demonstrated in this paper $F_i$ consist of direct pixel samples (\eg, taking a crop out of image $J_t$). In this case, \eqref{e:ftd} is a quadratic Least-Squares (LS) where each pixel of the minimizer $J$ is a weighted average of all its diffusion sample updates, \ie, 
\begin{equation}\label{e:closedform}
    \Psi(J_t|z) = \sum_{i=1}^n   \frac{F_i^{-1}(W_i)}{\sum_{j=1}^n F_j^{-1}(W_j)}\otimes F_i^{-1}(\Phi(I^i_t|y_i))
\end{equation}

\paragraph{Properties of MultiDiffusion.} The main motivation for the definition of $\Psi$ in \eqref{e:Psi} comes from the following observation: If we choose a probability distribution $P_\gJ$ such that 
\begin{equation}\label{e:initial_distribution_cond}
    F_i(J_T) \sim P_\gI, \qquad \forall i\in [n]
\end{equation}

and compute $J_{t-1}=\Psi(J_t|z)$, as defined in \eqref{e:Psi}, where we reach a zero FTC loss, $\gL_\ftd(J_{t-1}|J_{t},z)=0$, then: 
\begin{equation*}
    I^i_{t-1} = F_i(J_{t}) = \Phi( I^i_{t} | y_i)
\end{equation*}
That is, $I^i_t$, for all $i\in [n]$, is a diffusion sequence and thus $I^i_0$ is distributed according to the distribution defined by $\dif$ over the image space $\gI$. We summarize
\definecolor{mygray}{gray}{0.95}
\begin{center}\vspace{-10pt}			
    \colorbox{mygray} {		
      \begin{minipage}{0.977\linewidth} 	
       \centering
        \begin{proposition}
If $P_\gJ$ is a distribution over $\gJ$ satisfying \eqref{e:initial_distribution_cond}, 
and the FTD cost (\eqref{e:ftd}) is minimized to zero in \eqref{e:Psi} for all steps $T,T-1,\ldots,0$, then the images $I^i_t=F_i(J_{t})$ reproduce a $\dif$ diffusion path. In particular $F_i(J_0)$, $i\in [n]$ are distributed identically to samples from the reference diffusion model $\dif$. 
\end{proposition}
      \end{minipage}}			
\end{center}
The implications of this proposition are far reaching: using a single reference diffusion process we can flexibly adapt to different image generation scenarios without the need to retrain the model, while still being consistent with the reference diffusion model. Next, we instantiate this framework outlining several application of the Follow-the-Diffusion-Paths approach.

\begin{algorithm}[H]
    \caption{MultiDiffusion sampling.}\label{alg:FDS}
    \SetKwInOut{Input}{Input}
    \SetKwInOut{Output}{Output}
    \Input{ \; $\dif$~~~~~~~~~~$\triangleright$ pre-trained Diffusion Model \newline 
    \; $\{F_i\}_{i=1}^n$~~~$\triangleright$ image space mappings \newline
    \; $\{y_i\}_{i=1}^n$~~~~$\triangleright$ text-prompts conditioning  \newline
    \; $\{W_i\}_{i=1}^n$~~$\triangleright$ per-pixel weights
    }
    $J_T\sim P_\gJ$ ~~~$\triangleright$ noise initialization \newline
    \For{$t=T,...,1$}{
     \; $I_{t-1}^i \gets \Phi(F_i(J_t),y_i) \;\;\forall i\in [n]$ ~~~$\triangleright$ diffusion updates \newline
    $J_{t-1}\gets\texttt{MultiDiffuser}(\set{I^i_{t-1}}_{i=1}^n)$ ~~~~~~~~~ $\triangleright$ \eqref{e:closedform}
    }
    \Output{ $J_0$}
\end{algorithm}

\section{Applications}\label{sec:applications}
\subsection{Panorama}\label{sec:panorama}\hspace{-10pt}
As a first instantiation we use our framework to define a diffusion model in an image space $\gJ$ with $H'\geq H$, $W'\geq H$ directly from a trained model $\dif$ working in image space $\gI$. Let $\gZ=\gY$ (namely, generating a panoramic image for a given text-prompt), $F_i(J)\in \gI$ is an $H\times W$ crop of image $J$, and $z=\lambda_i(z)$. We consider $n$ such crops that cover the original images $J$. Setting $W_i=\mathbf{1}$, we get  
\begin{align}
\label{eq:panorama}
    \Psi(J_t,z)  = &\argmin_{J\in \gJ}\ \  \sum_{i=1}^n \norm{F_i(J) - \Phi(F_i(J),z) }^2 
\end{align}
that is a least-squares problem, the solution of which is calculated analytically according to \eqref{e:closedform}. See the Appendix \ref{appendix:panorama} for implementation details.

\begin{figure} [t!]
    \centering
    \begin{tabular}[width=\linewidth]{c}
    \includegraphics[width=0.96\linewidth]{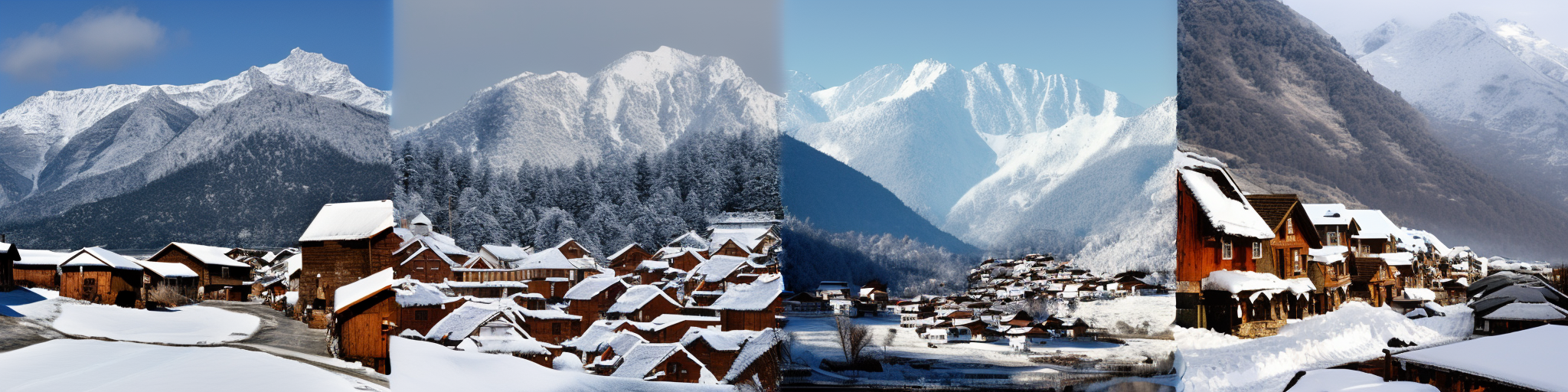}  \\
    \small{(a) Generation with per-crop independent diffusion paths.} \\
        \includegraphics[width=0.96\linewidth]{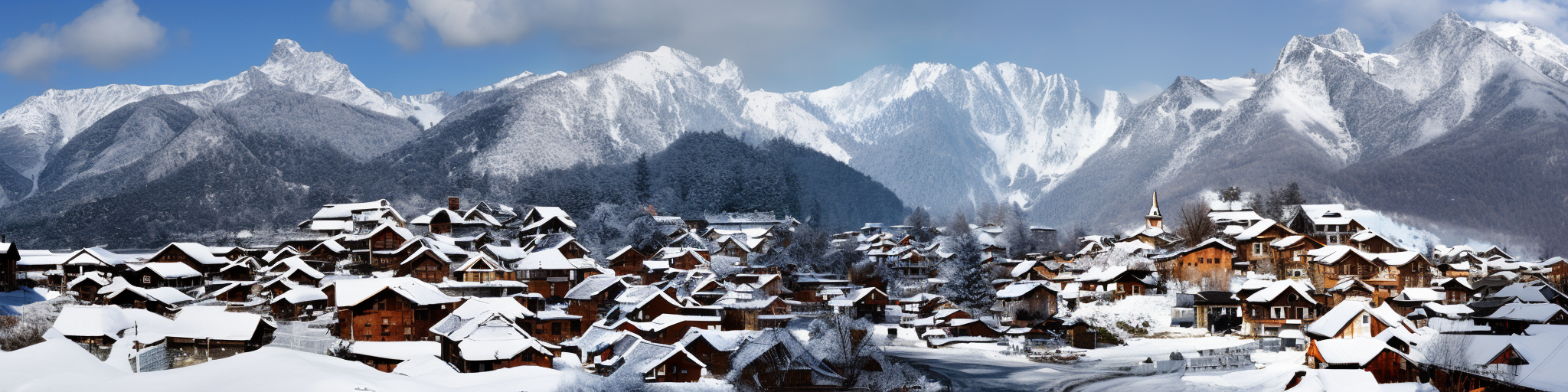} \hspace{-5pt} \\
        \small{(b) Generation with fused diffusion paths using MultiDiffusion.}
     \end{tabular}
    \caption{Independent diffusion paths vs. MultiDiffusion. (a) Panoramic image generated by applying the reference model on four crops independently; as expected, there is no coherency between the crops. (b) Starting from the same noise, our generation process steers these initial diffusion paths into a consistent and high quality image.}
    \label{fig:pano_path}
\end{figure}
As discussed in Sec.~\ref{sec:method}, \emph{MultiDiffusion} reconciles multiple diffusion paths provided by the reference model $\Phi$. We illustrate this property  in Fig.~\ref{fig:pano_path}, where we consider a panorama of $H \times 4W$.  Fig.~\ref{fig:pano_path}(a) shows the generation result when independently applying $\dif$ on four non-overlapping crops. As expected, there is no coherency between the crops since this amounts to four random samples from the model. Starting from the same initial noise, our generation process (Eq.~\ref{eq:panorama}), allows us to fuse these initially-unrelated diffusion paths, and steer the generation into a high-quality, coherent panorama (b).


\subsection{Region-based text-to-image-generation}
\label{sec:region}

Given a set of region-masks $\{M_i\}_{i=1}^n \subset \{0,1\}^{H\times W}$ and a corresponding set of text-prompts  $\{y_i\}_{i=1}^n \subset \gY^n$, our goal is to generate a high-quality image $I \in \gI$ that depicts the desired content in each region. That is, the image segment $I \otimes M_i$ should manifest $y_i$. Going back to our formulation (\eqref{e:Psi_series}), the \emph{MultiDiffusion} process is defined over the condition space $\gZ=\gY^n$, \ie, $z=(y_1,\ldots,y_n)$, and the target image space $\gJ=\gI$ is identical to the reference one:
$$\Psi:\gI\times \gY^n \too \gI$$
Furthermore, the region selection maps are defined as $F_i(I)=I$, the pixel weights are set according to the masks, $W_i=M_i$, and the $\Psi$ step is defined as the solution to the least-squares problem: 
\begin{align}
    \Psi(J_t,z)  = &\argmin_{J\in \gI}\ \ \sum_{i=1}^n \Big \| M_i \otimes \Big [J - \Phi(J_t|y_i) \Big ] \Big \|^2
\end{align}
\label{eq:region_rough}
The solution to this LS problem is calculated analytically. At each step we apply the pretrained diffusion w.r.t. each of the given prompts, resulting in multiple diffusion directions $\dif(J_t|y_i)$. We encourage each pixel in $J_t$ to follow the (averaged) directions associated with the regions $M_i$ containing it (Eq.~\ref{e:closedform}).
\paragraph{Fidelity to tight masks} \label{sec:bootstrapping} We further support obtaining high-fidelity to tight masks if provided by the user (see Fig. \ref{fig:region}).
  We noticed that the layout is being determined early on in the diffusion process, and thus we strive to encourage $\dif(J_t|y_i)$ to focus on the region $M_i$ early on in the process in order to match the desired layout, and to consider the full context in the image next, to achieve an harmonized result. We integrate time dependency in the maps $F_i$, introducing a bootstrapping phase. That is, 
  \begin{equation}\label{e:bootstrap}
 F_i(J_t, t)=\begin{cases}
			J_t, & \text{if $t\leq T_{init}$}\\
                 M_i \otimes J_t + (1-M_i) \otimes S_t, & \text{otherwise}
		 \end{cases}     
  \end{equation}
Where $T_{init}$ is the bootstrapping stopping step parameter, and $S_t$ is a random image with a constant color, which serves as background (see Appendix \ref{appendix:bootstrapping} for implementation details).

We demonstrate the efficiency of our bootstrapping approach in Sec.~\ref{sec:region_results}. We set $T_{init}$ to be $20\%$ of the generation process (\ie,  $T_{init}=800$).

\begin{figure*} [t!]
    \centering
    \includegraphics[width=0.95\textwidth]{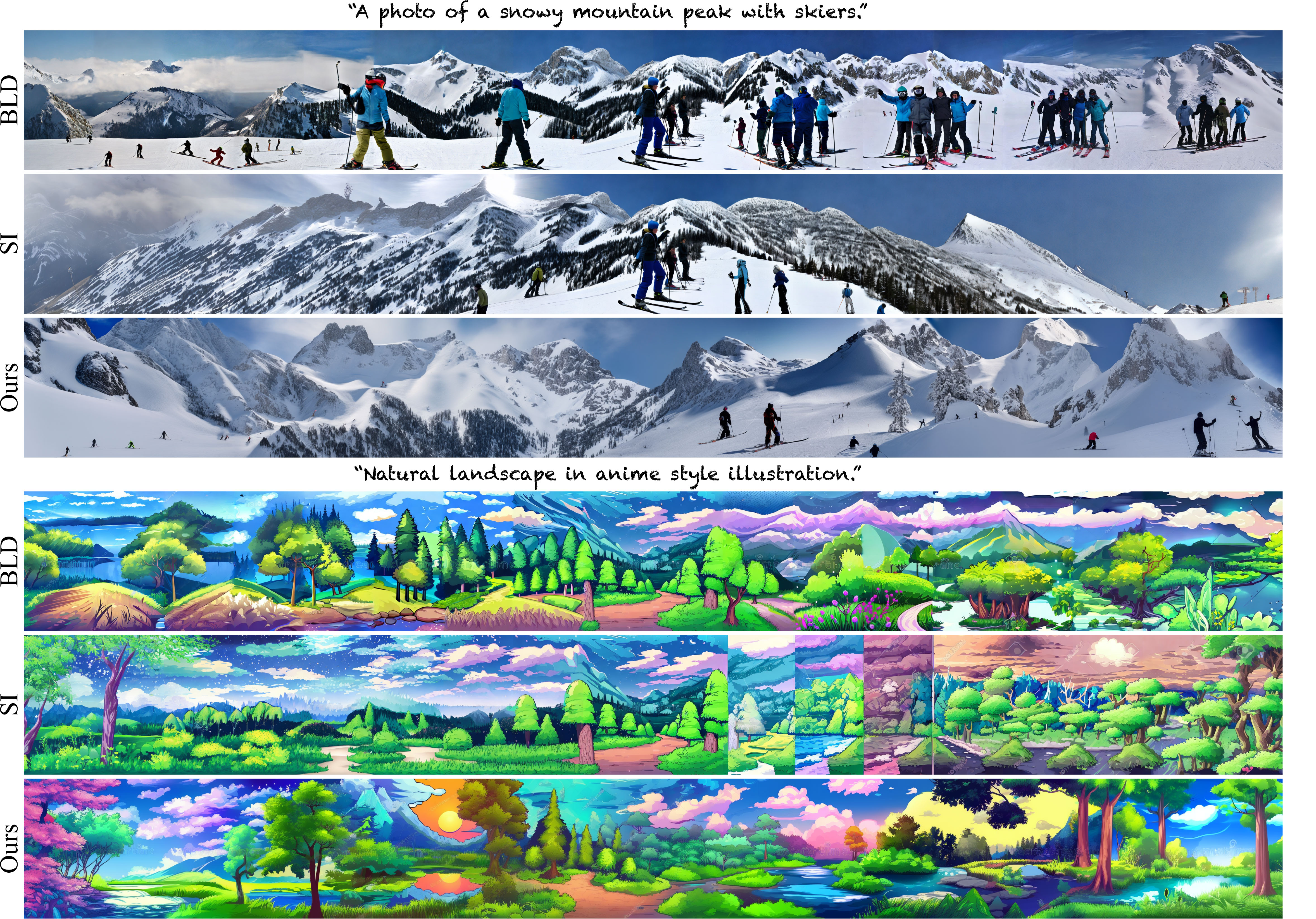}\hspace{-10pt}
    \caption{Text-to-Panorama comparison to Blended Latent Diffusion (BLD) \cite{avrahami2022blended} and Stable Inpainting (SI) \cite{rombach2022high}. Our framework produces seamless and diverse content whereas the baselines either contain repetitive content, visible seams or artifacts.}
    \label{fig:pano}
\end{figure*}

\section{Results}
\label{sec:results}
We thoroughly evaluate our method when applied to each task as discussed in Sec.~\ref{sec:applications}. In all experiments, we used Stable Diffusion \cite{rombach2022high}, where the diffusion process is defined over a latent space $\gI = \Real^{64\times 64 \times 4}$, and a decoder is trained to reconstruct natural images in higher resolution $[0,1]^{512\times 512 \times 3}$. Similarly, the MultiDiffusion process, $\Psi$ is defined in the latent space $\gJ = \Real^{H`\times W` \times 4}$ and using the decoder we produce the results in the target image space $[0,1]^{8H'\times 8W' \times 3}$. 
We use the public implementation of Stable Diffusion by HuggingFace \cite{von-platen-etal-2022-diffusers}, with the v2 pre-trained model.

\subsection{Panorama Generation}\label{sec:panorama_results}
To evaluate our method on the task of text-to-panorama generation (Sec.~\ref{sec:panorama}),   we generated a diverse set of $512 \times 4608$ panoramas, $\times 9$ wider than the original training resolution. Since there is no direct method for generating images at arbitrary aspect ratio from text, we compare to the following two baselines: (i) Blended Latent Diffusion (BLD) \cite{avrahami2022blended} (combined with Stable Diffusion \cite{rombach2022high}), and Stable Inpainting (SI) \cite{rombach2022high}, which has been finetuned on large-scale data for inpainting. For both baselines, the panoramic image is generated gradually, starting from a central image (sampled by $\Phi$ given the input text), and extrapolated 
progressively to the right and left. 

Fig.~\ref{fig:pano} shows sample generation results by our method compared to the above baselines. As seen, both baselines often exhibit visible seams and discontinuities between overlapping crops, as well as degradation in visual quality as moving away from the center pivotal image; this is expected due to the iterative generation process. BLD often generates repetitive content (e.g., skiers example), where SI results in noticeable visual difference between the left and right parts of the image. In contrast, our framework  \emph{simultaneously} ``samples'' the panoramic image by combining the diffusion paths of \emph{all} crops, resulting in seamless and high quality images.
 Additional comparisons are in the Appendix \ref{fig:pano_sm}.

\begin{table}[h]
\renewcommand{\tabcolsep}{1.0pt}
\begin{tabular}{@{}l|ccc}
& FID $\downarrow$&CLIP-score$ \uparrow$&CLIP-aesthetic$ \uparrow$ \\ \hline
Stable Diffusion & $6.05 \pm 3.1$ & $0.27$ & $6.36$ \\ \hdashline 
SI & $45.5 \pm 14.5$ & $0.26$ & $5.76$ \\
BLD & $18.4 \pm 7.4$ & $\bf{0.27}$ & $6.02 $ \\
Ours        &   $\bf{10.3 	\pm 4.8} $  &  $\bf{0.27 }$   & $\bf{6.36}$   \\
\end{tabular}
\caption{
Panorama generation evaluation. We report FID, CLIP text-image score, and CLIP aesthetic scores for of our method compared to the baselines. See more details in Section.~\ref{sec:panorama_results}.}
\label{tab:panorama}
\end{table}


To quantify these observations, we use the Frechet Inception Distance (FID) \cite{parmar2021cleanfid} to measure the distance between the distribution of $512 \times 512$ crops from the panoramic images to the distribution of images generated by the reference model $\dif$. That is, for a given text prompt, we sample $N$ different $512 \times 512$ images from $\dif$, and consider them as our reference dataset. For the baselines and our method, we generated $N$ panoramic images, and randomly sampled a $512 \times 512$ crop from each sample to serves as the generated dataset and computed the FID accordingly.  

To further assess  the quality of our results, we evaluated two CLIP-based scores: (i) text-image CLIP score \cite{radford2021learning} measured by the cosine similarity between the text prompt and the image embeddings, and (ii) CLIP aesthetic \cite{Schuhmann2022LAION5BAO} measured by a linear estimator on top of CLIP predicting the aesthetic quality of the images. 

We used  $N=2000$ samples and repeated this evaluation for $8$ different text-conditioning.  Table~\ref{tab:panorama} reports the mean and standard-deviation of FID and CLIP scores for our method and the baselines. We additionally report the scores for an  independent set of samples images from $\dif$, which serves as a baseline. As seen, our method outperforms the existing baselines in all metrics.

\begin{figure} [t!]
    \centering
    \includegraphics[width=\linewidth]{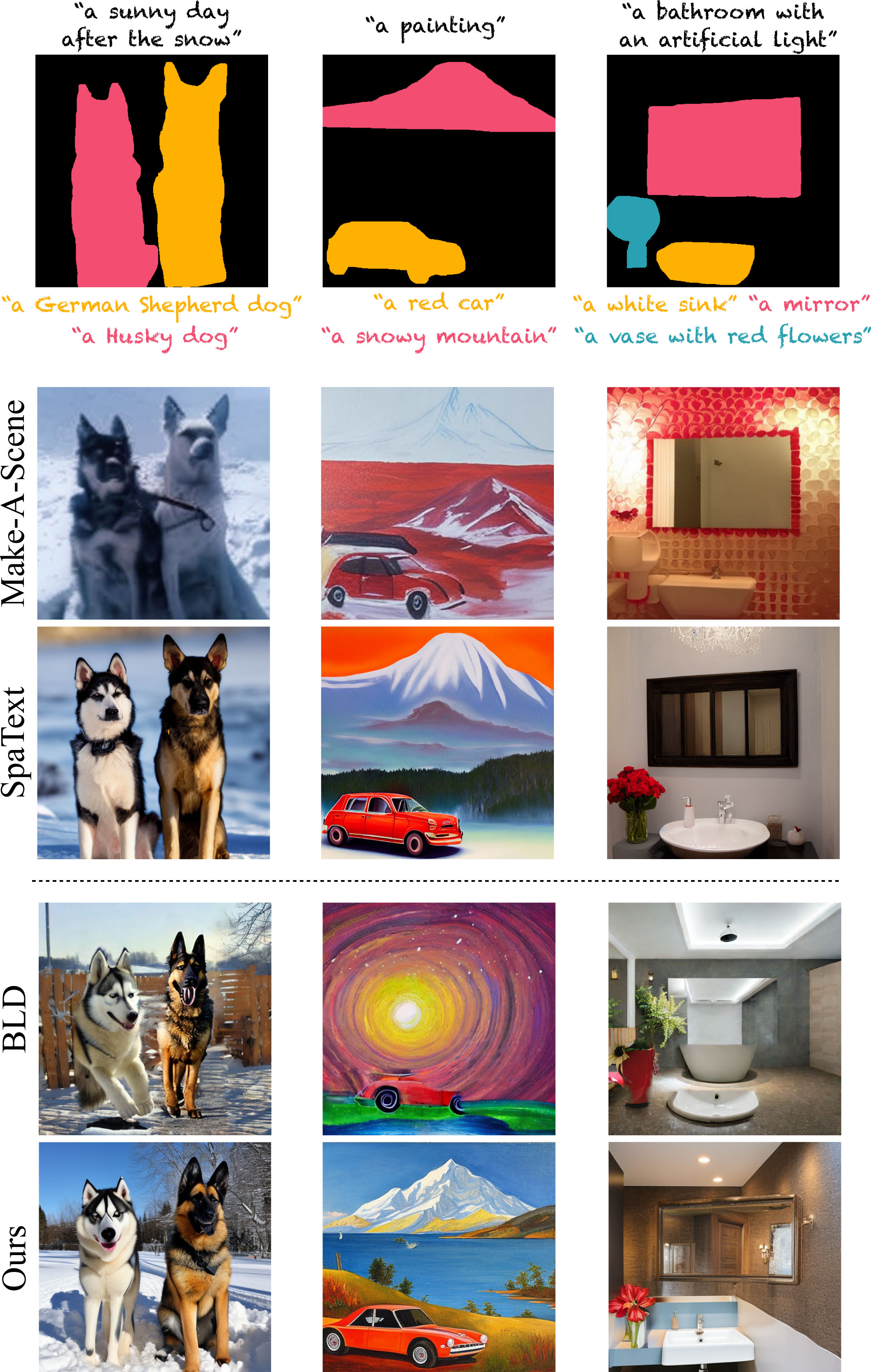}\hspace{-0pt}
    \caption{Region-based text-to-image generation. The input segmentation maps with the corresponding region text  descriptions are shown above each example. Below: Make-A-Scene \cite{gafni2022_make_a_scene}and SpaText \cite{avrahami2022spatext} -- trained specifically for this task on a large-scale segmentation-text-image dataset;  Blended Latent Diffusion (BLD) \cite{avrahami2022blended}, and our results.}
    \label{fig:region}
\end{figure}

\subsection{Region-based Text-to-Image Generation}\label{sec:region_results}

Our region-based formulation (Sec.~\ref{sec:region}) allows novice users greater flexibility in their content creation, by lifting the burden of creating accurate tight masks. As can be seen in Fig.~\ref{fig:teaser}, Fig.~\ref{fig:rough_variability} and Fig.~\ref{fig:rough1}, our method generates diverse high-quality samples that comply with text description, given only bounding boxes region guidance. As seen in Fig.~\ref{fig:rough_variability}, by starting our generation from a different input noise, we can generate diverse samples, depicting objects in different scales and appearances, all following the same spatial controls. Notably, since we integrate the controls from all regions into a unified generation process, our method can generate complex scene effects (e.g.,  background blur, shadows or reflections) which are coherently immerse in the scene.  More results are included in the Appendix.

%
We compare our region-based framework with Make-A-Scene \cite{gafni2022_make_a_scene} and the concurrent work SpaText \cite{avrahami2022spatext}. 
Both baselines perform large-scale training specifically for this task. Note that these models are not publicly available, thus we qualitatively compare to their provided examples. 

Additionally, we consider an adaptation of BLD \cite{avrahami2022blended} as a baseline. Similarly to Sec.~\ref{sec:panorama_results}, this is done by applying their method in an auto-regressive manner by first generating the background, and sequentially generating each of the foreground objects.  

As seen in Fig.~\ref{fig:region}, our framework produces consistent images that adhere to the spatial constraints, and are qualitatively on par with \cite{avrahami2022spatext}. The auto-regressive approach based on BLD \cite{avrahami2022blended} often results in incoherent images and  an unnatural scene.  (e.g., misplaced sink in   ``bathroom'' example). Additional comparisons to the baselines are in the Appendix.

\begin{table}[]
    \centering
  \begin{tabular}{l|c}
& IoU $\uparrow$ \\ \hline
COCO dataset & $0.43 \pm 0.09$ \\ \hdashline 
SI & $0.16 \pm 0.10$ \\
BLD  & $0.17 \pm 0.11$ \\
Ours \small{\textit{w/o bootstrapping}} & $0.18 \pm 0.10$\\
Ours & $\bf{0.26 \pm 0.12}$ \\ 
\end{tabular}
\caption{\small{Region-based generation evaluation of the COCO dataset. We evaluate Intersection over Union (IoU), see Sec. \ref{sec:region_results} for details.}}
\label{tab:region}  
\end{table}

\begin{figure}[b]
    \centering
    \includegraphics[width=\linewidth]{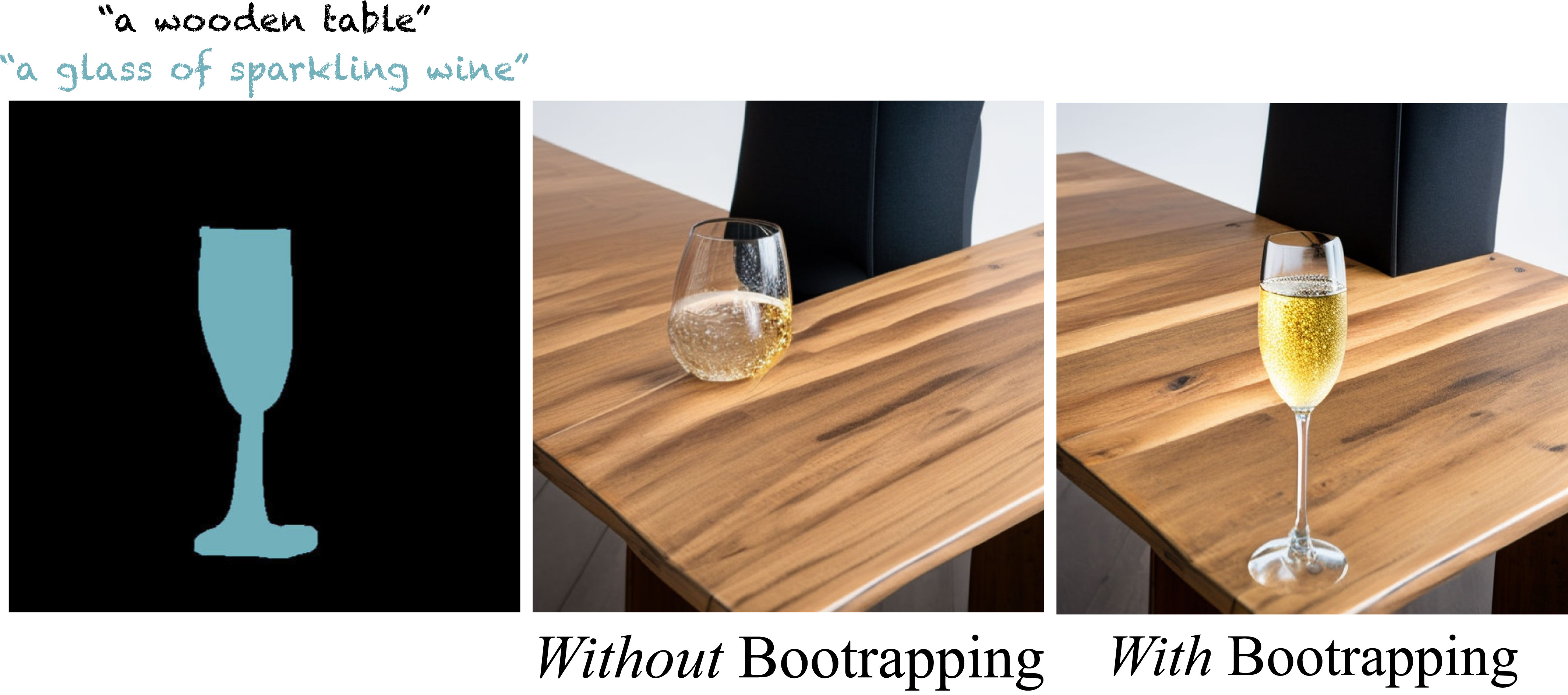}
    \caption{Bootstrapping ablation. Without bootstrapping (middle), our method successfully generates the glass in \emph{some} location inside the mask (left). With our bootstrapping mechanism (right), we achieve high-fidelity to the provided tight mask. See Sec~\ref{sec:bootstrapping} for details.}
    \label{fig:bootstrap}
\end{figure}

%
%
To quantitatively evaluate our performance, we use the COCO dataset \cite{lin2014microsoft}, which contains images with global text caption and instance masks for each object in the image. 
We apply our method on a subset from the validation set,  obtained by filtering examples which consists of 2 to 4 foreground objects, excluding people, and masks that occupy less than $5\%$ of the image. This results in $1K$ diverse samples. Following \cite{avrahami2022spatext}, we use the ground truth labels to provide a text prompt for each foreground region, \ie, ``a \{label\}'', and use the full image caption as the prompt describing the background. 

We evaluate the results with an off-the-shelf segmentation model \cite{cheng2022masked} on the generated images, and measure the Intersection over Union (IoU) w.r.t.~to the ground-truth segmentation. Table~\ref{tab:region} reports the performance for our method and the baselines described above. As an upper bound, we also report the IoU w.r.t. the original images in the set. Note that our method outperforms the existing baselines SI \cite{rombach2022high} and BLD \cite{avrahami2022blended}. We additional provide qualitative examples are included in the Appendix.

Finally, we present an ablation of our bootstrapping stage ( \eqref{e:bootstrap}): qualitatively in Fig.~\ref{fig:bootstrap}, and quantitatively in Table \ref{tab:region}. Note that without bootstrapping, our framework still generates the desired object within the mask region, however, the bootstrapping stage makes it tighter to the given mask.

\begin{figure*} [t!]
    \centering
    \includegraphics[width=0.95\textwidth]{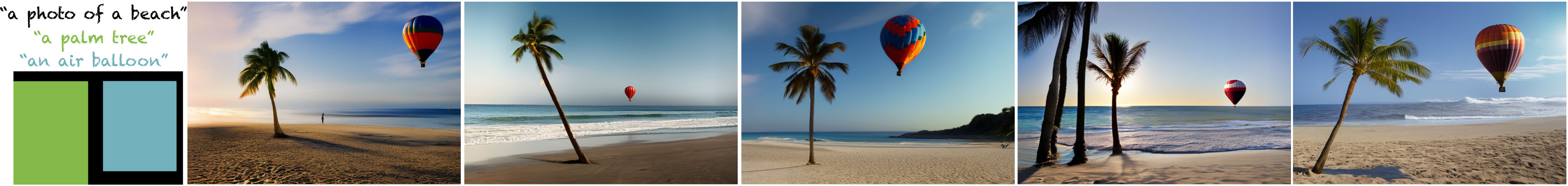}\hspace{-10pt}
    \caption{Diverse samples generated by our framework, given rough scene layout guidance (left). All images depict sensible composition, scene effects and relative size of objects.}
    \label{fig:rough_variability}
\end{figure*}

\begin{figure*} [t!]
    \centering
    \includegraphics[width=0.95\textwidth]{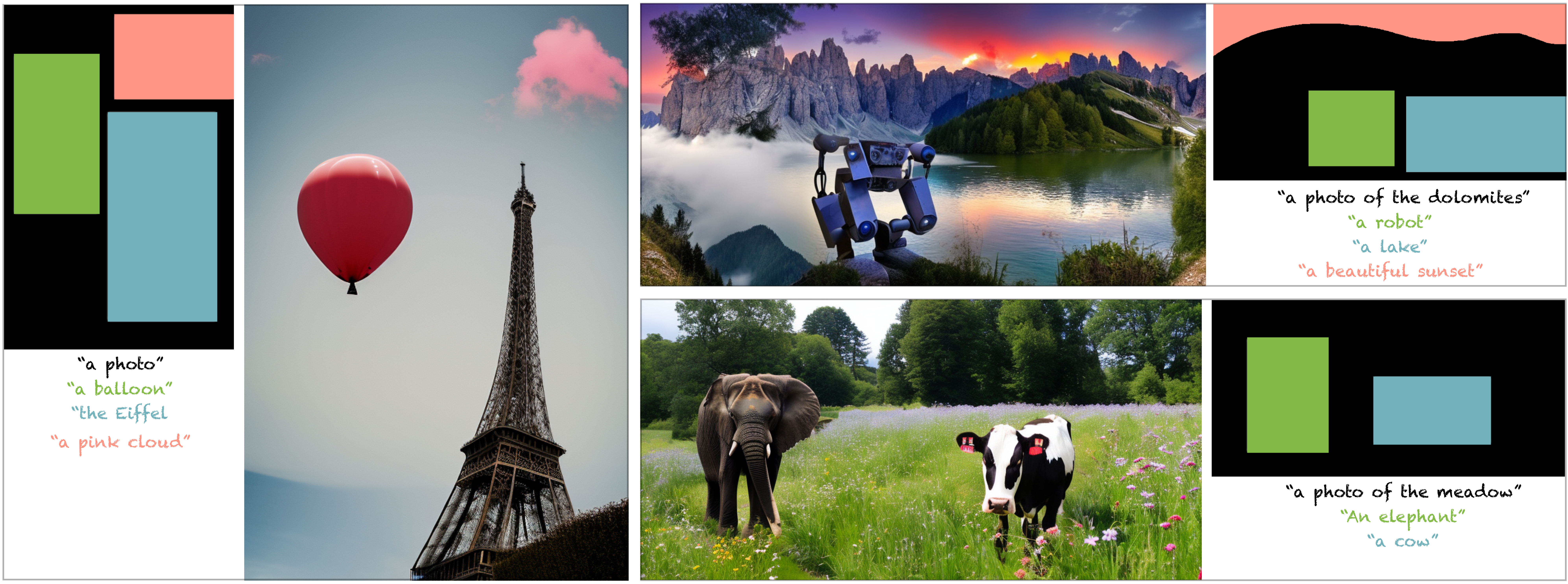}\hspace{-10pt}
    \caption{Rough masks. Sample results of our region-based generation approach (see Sec.~\ref{sec:region}). Our method can work with rough masks, the can intuitively be obtained by novice users. 
    }
    \label{fig:rough1}
\end{figure*}

\section{Discussion and Conclusions}
\label{sec:conclusions}
\begin{SCfigure}
    \includegraphics[width=0.56\linewidth]{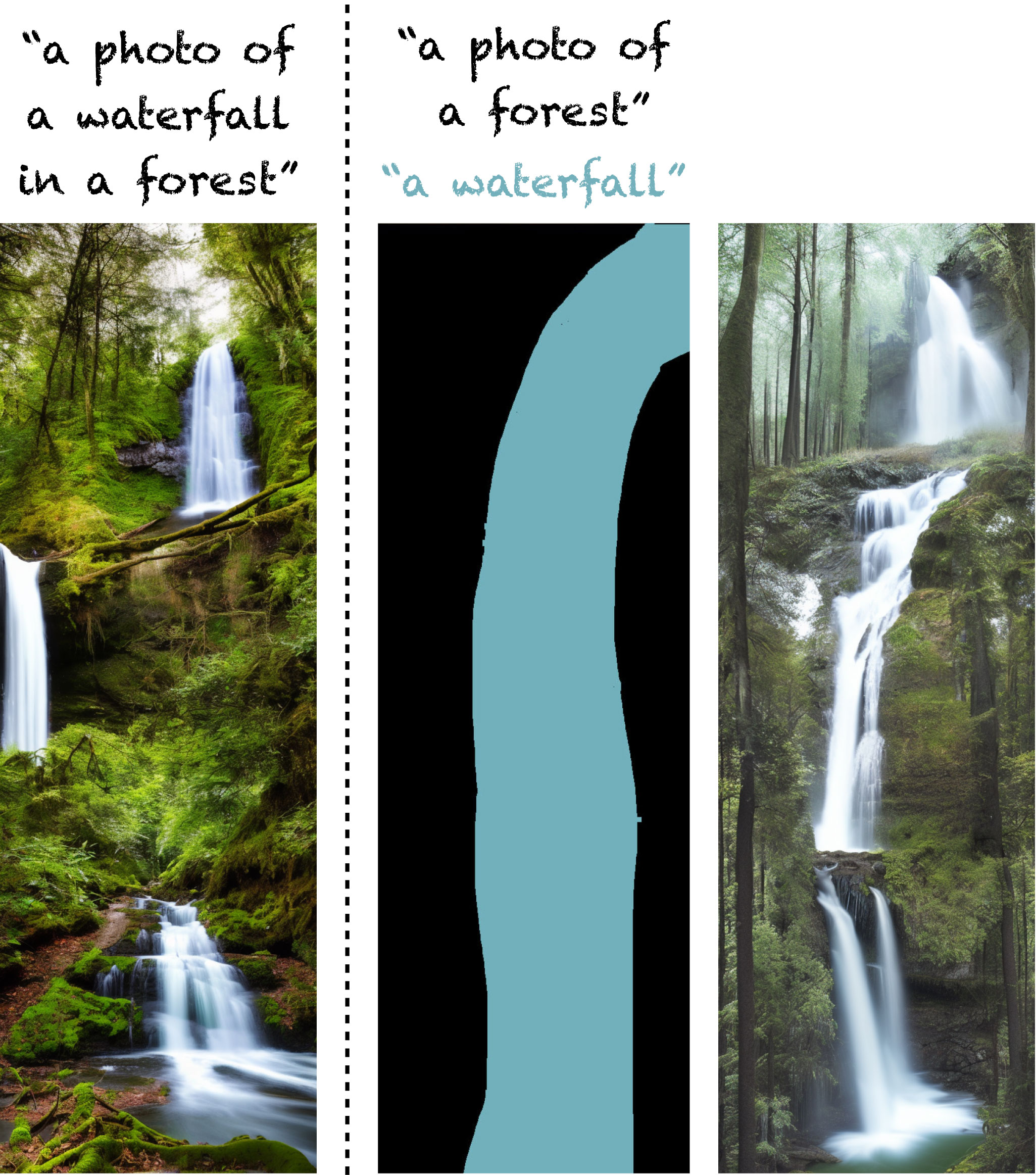}
    \caption{\protect\rule{0ex}{0ex}Our method heavily relies on the prior of the reference diffusion model. Left: our method when applied to a vertical panorama. The reference diffusion model is biased towards adding a waterfall in each viewing crop, resulting with an unnatural scene. Right: we can try to overcome this by adding a region-based constraint.
    }\hspace{-7pt}
    \label{fig:limitation}
\end{SCfigure}

Controllable generation is one of the major pending challenges with text-to-image diffusion models. We proposed to tackle this challenge from a fundamentally new direction -- defining a new generation process on top of a pre-trained and fixed diffusion model. This approach has several key advantages over previous works: (i) it does not require any further training or finetuning, (ii) it can be applied to various different generation tasks, and (iii) our generation process yields an optimization task which can be solved in closed form for many tasks, hence can be computed efficiently, while ensuring convergence to the global optimum of our objective.
As for limitations, our method heavily relies on the generative prior of the reference diffusion model, \ie, the quality of our results depends on the diffusion paths provided by the model. Thus, when a ``bad'' path is chosen by the reference model (\eg, bad seed, or biased text-prompt), our results will be affected as well. In some cases, we can mitigate it by introducing more constraints into our framework (bootstrapping in Sec.~\ref{sec:region}), or prompt-engineering (Fig.~\ref{fig:limitation}). 
We thoroughly evaluated our framework, demonstrating state-of-the-art results even compared to methods that are tailored-trained for specific
tasks. 

We believe that our work can trigger further future research in harnessing the power of a pre-trained diffusion model in more principled manner. One way forward, for example, is to generalize the MultiDiffusion with a more general optimization problem, 
 \begin{equation}
    \begin{aligned}
    \Psi(J_t|z)  = &\argmin_{J\in \gJ}\ \  \gL_{\ftd}(J|J_t,z) + \gL_0(J,J_t,z)\\
     & \qquad \ \text{s.t.} \ \ \   J\in \gC(J_t,z) 
    \end{aligned} 
\end{equation}
where $\gL_0$ is a cost function and $\gC$ is a set of (hard) constrains that control the MultiDiffusion process by incorporating other priors and/or design constraints. This approach provides a further of freedom in designing MultiDiffusion processes. 


\section{Acknowledgments}

LY is supported by a grant from Israel CHE Program for Data Science Research Centers and the Minerva Stiftung. OB is supported by the Israeli Science Foundation (grant 2303/20).
The research was supported also in part by a research grant from the Carolito Stiftung (WAIC). We thank Michal Geyer and Dolev Ofri-Amar for proofreading the paper.


\bibliography{example_paper}
\bibliographystyle{icml2023}

\newpage
\appendix
\onecolumn

\section{Additional Results}
In the following section we provide additional results and comparisons for the applications shown in the main paper.


\subsection{Panorama Generation}
\begin{figure*} [t!]
    \centering
    \includegraphics[width=0.9\textwidth]{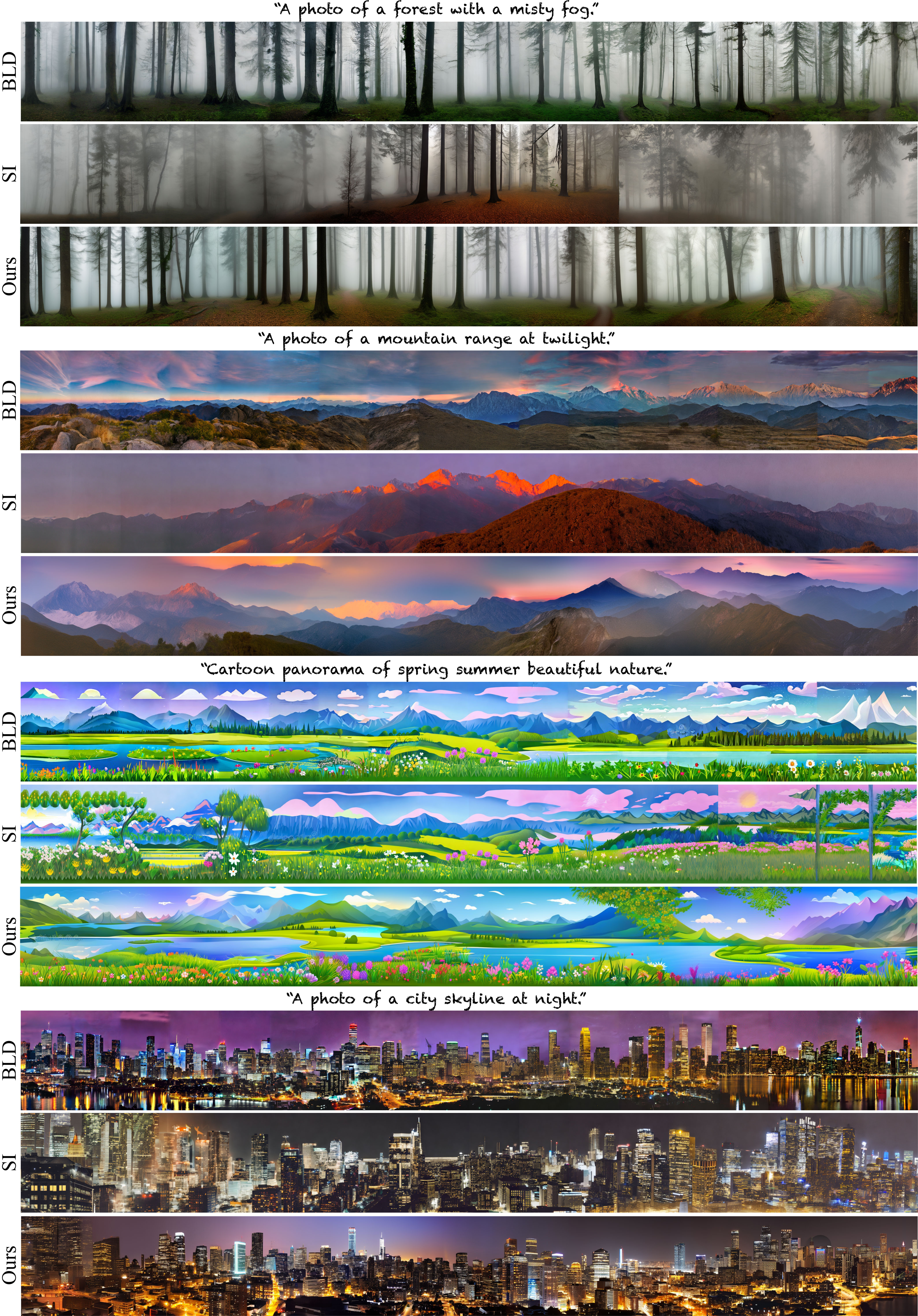}\hspace{-10pt}
    \caption{Text-to-Panorama additional results and comparisons to Sec.~\ref{sec:panorama_results}. 
    }
    \label{fig:pano_sm}
\end{figure*}

We provide additional results and qualitative comparisons for the task of text-to-panorama (Sec.~ \ref{sec:panorama_results}). Fig.~\ref{fig:pano_sm} depicts additional comparisons of our method vs Stable Inpainting (SI) \cite{rombach2022high} and Blended Latent Diffusion  (BLD) \cite{avrahami2022blended}. We also show vertical panorama result in Fig.~\ref{fig:rough_sm_2} left.

\subsection{Region-based Text-to-Image Generation}
We provide additional qualitative results and comparisons for the task of region-based generation (Sec.\ref{sec:region}) in  Fig.~\ref{fig:rough_sm_2} and Fig.~\ref{fig:region_based_sm}.

\subsection{Region-based Text-to-Image Generation on COCO}
We include sample results and comparison on the subset from the validation set of COCO in Fig.~\ref{fig:coco_cm}. See more details about this experiment in Sec.~\ref{sec:region_results}.

\begin{figure*}[t!]
    \centering
    \includegraphics[width=\textwidth]{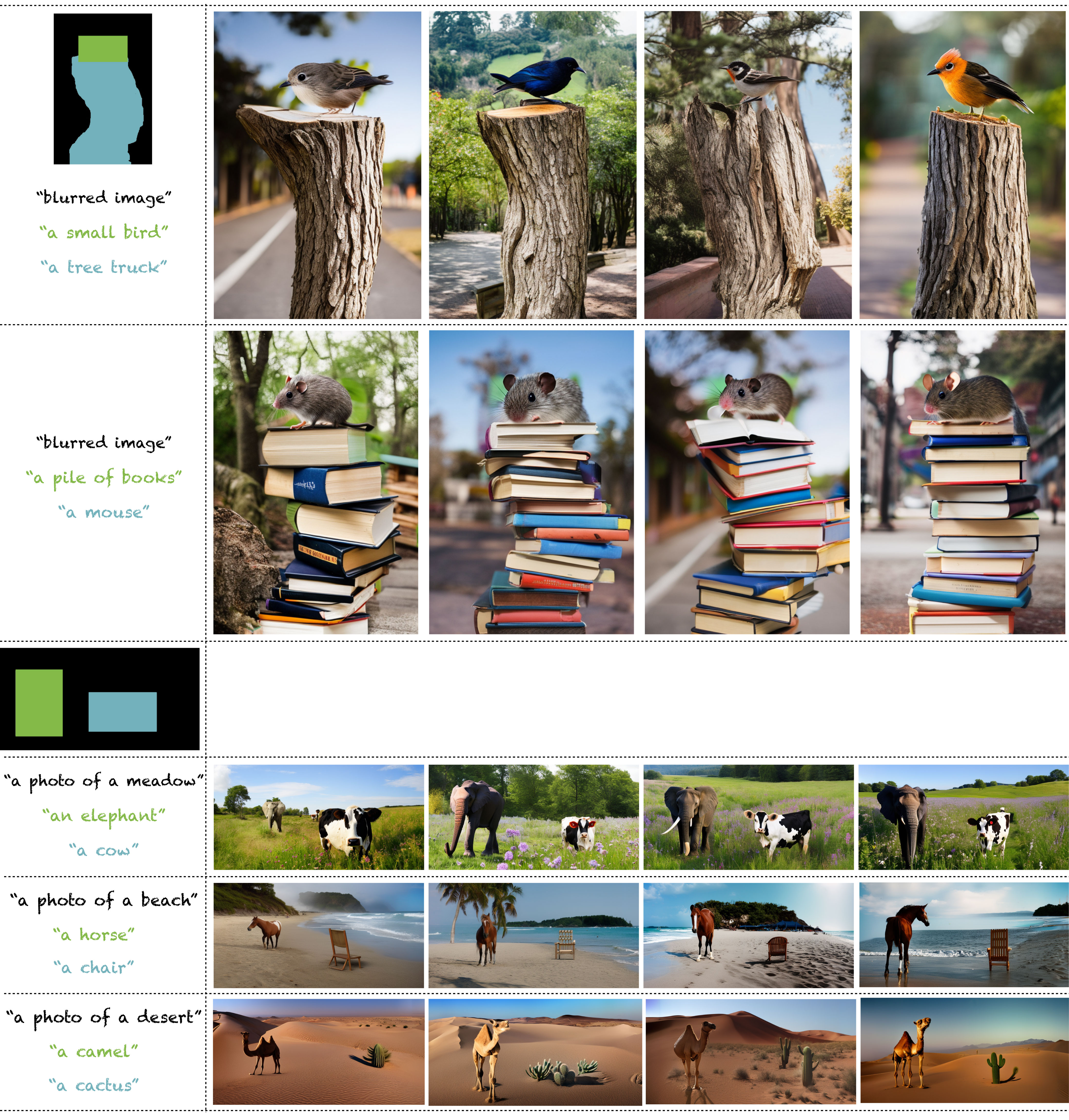}\hspace{-10pt}
    \caption{Additional results of our method on generation from rough scene layouts (~Sec.~\ref{sec:region_results}). For each spatial layout, and for each text prompt, we show different samples from our method. 
    }
    \label{fig:rough_sm}
\end{figure*}

\begin{figure*}[t!]
    \centering
    \includegraphics[width=\textwidth]{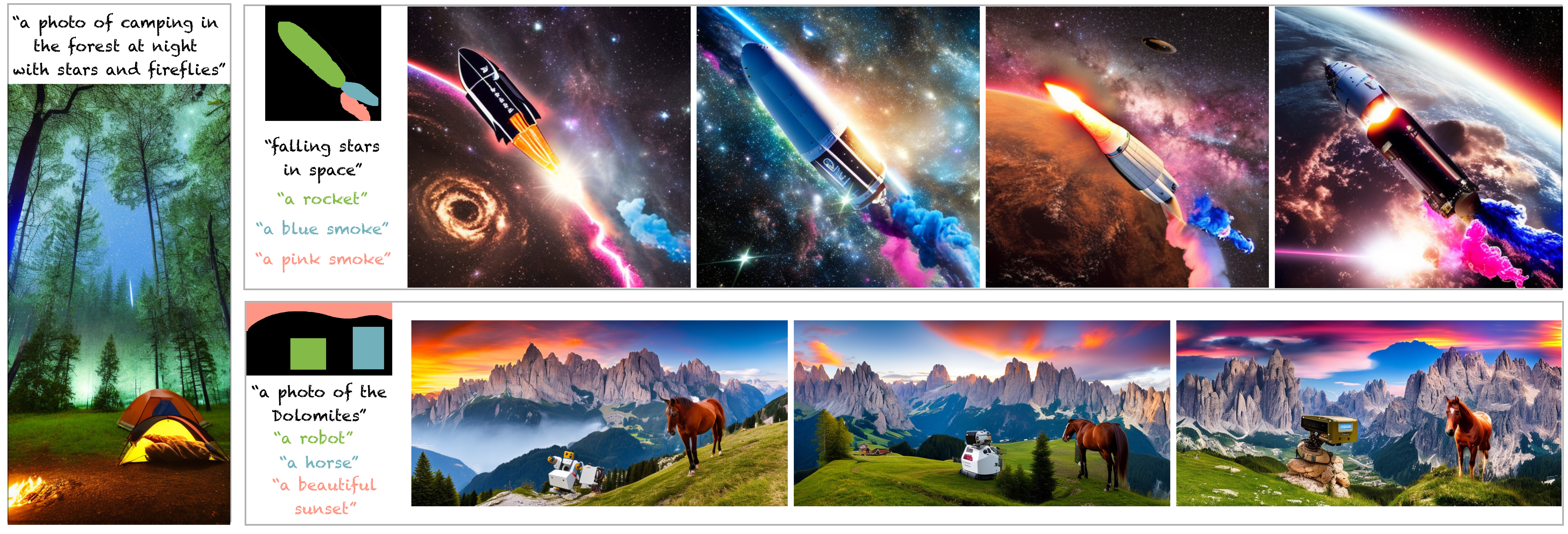}\hspace{-10pt}
    \caption{Left: vertical panoramic image ($1024 \times 512$) generated by our method. Right: additional generation results using a combination of rough and tight regions; for each layout, we present diverse generated samples.}
    \label{fig:rough_sm_2}
\end{figure*}

\begin{figure*}[t!] 
    \centering
    \includegraphics[width=\textwidth]{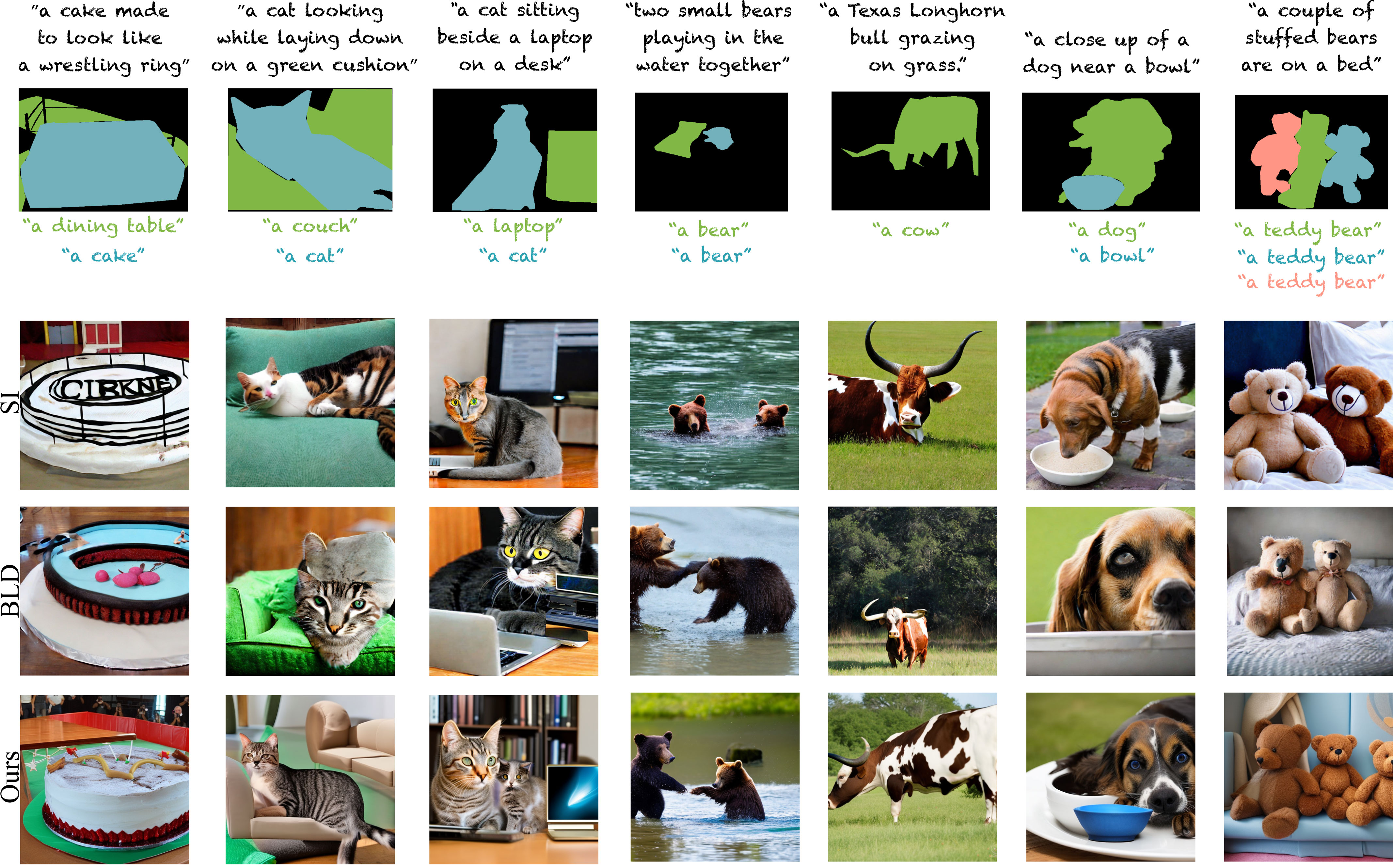}\hspace{-10pt}
    \caption{Sample results from COCO validation set by BLD, SI and our method. See more details in Sec.~\ref{sec:region_results}. 
    }
    \label{fig:coco_cm}
\end{figure*}

\begin{figure*}[t!]
    \centering
    \includegraphics[width=0.8\textwidth]{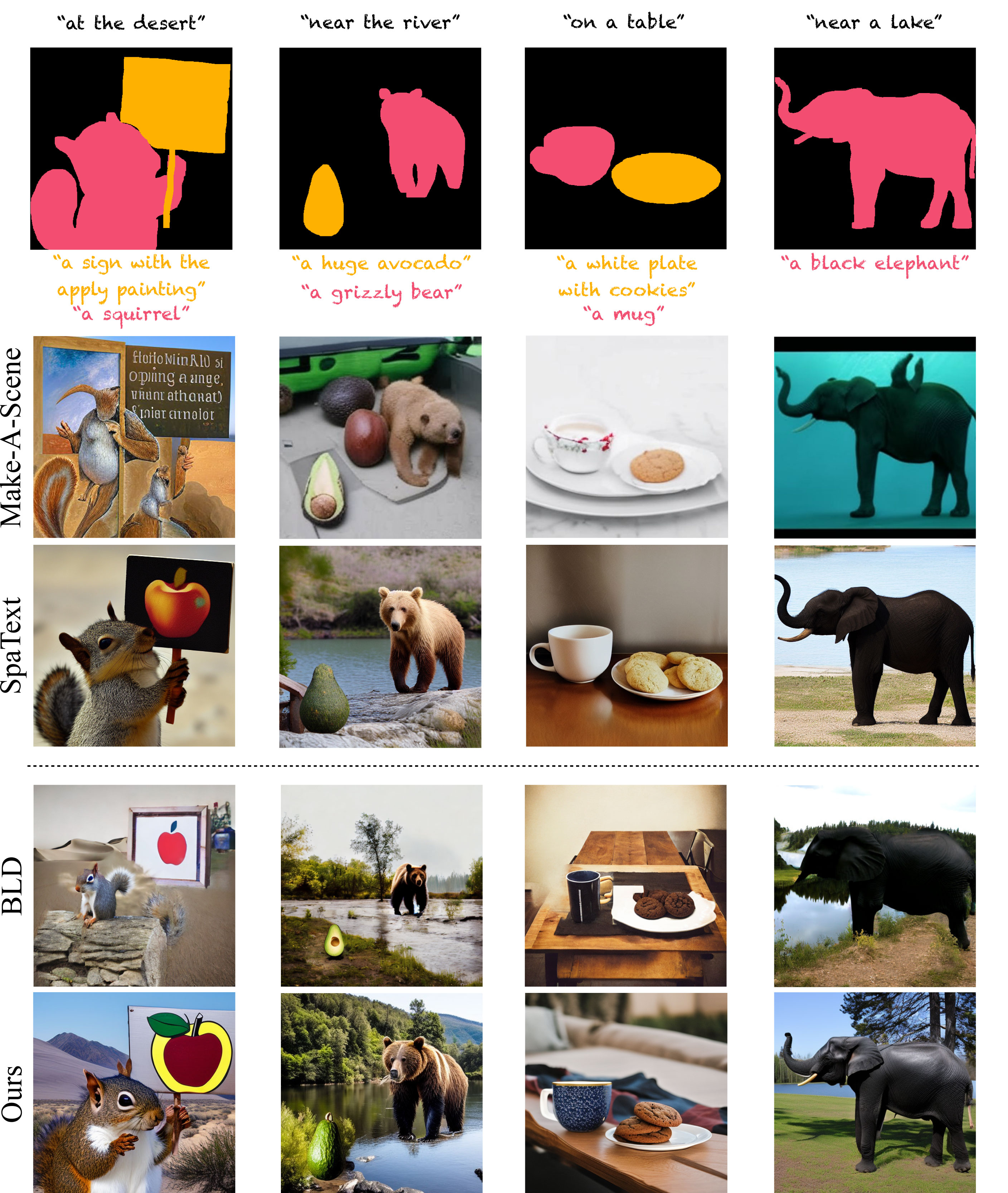}\hspace{-10pt}
    \caption{Additional qualitatively comparison to Make-A-Scene \cite{gafni2022_make_a_scene}, Blended-Latent-Diffusion \cite{avrahami2022blended}, Spa-Text \cite{avrahami2022spatext}, and ours. See Sec.~\ref{sec:region}, and Sec.~\ref{sec:region_results}
    }
    \label{fig:region_based_sm}
\end{figure*}

\section{Additional Implementation Details.}
\subsection{Panorama (Sec.~\ref{sec:panorama})}
\label{appendix:panorama}
In the case of panorama generation, our maps $F_i$ are defined as fixed-size crops from the full panorama. Specifically, for a panorama with spatial resolution $H'\times W'$, we consider overlapping crops of size $H\times W$ where $H=W=64$ defined in the Stable Diffusion latent space (which translates to size $512\times 512$ in RGB space). Our maps $F_i,...,F_n$ provide crops with a sliding window of size $\texttt{step}=8$ in the latent space (64 pixels in RGB space). In particular, $n={\frac{H'-64}{\texttt{step}} \cdot {\frac{W'-64}{\texttt{step}}}}$. \newline We summarize, 


\begin{algorithm}[H]
    \caption{MultiDiffusion sampling - Panorama.}\label{alg:FDS}
    \SetKwInOut{Input}{Input}
    \SetKwInOut{Output}{Output}
    \Input{ \; $\dif$~~~~~~~~~~$\triangleright$ pre-trained Diffusion Model \newline 
    \; $H', W'$~~~$\triangleright$ resolution of the desired panorama \newline
    \; $\{F_i\}_{i=1}^n$~~~$\triangleright$ mappings defining crops from the panorama  \newline
    \; $y$~~~$\triangleright$ conditioned text-prompt
    }
    $J_T\sim \mathcal{N}(0,I) \;\; J_T\in \mathcal{R}^{H'\times W' \times C}$ ~~~$\triangleright$ noise initialization \newline
    \For{$t=T,...,1$}{
     \; $I_t^i \gets F_i(J_t) \;\;\forall i\in [n]$ ~~~$\triangleright$ take crops from the panorama \newline
     $I_{t-1}^i \gets \Phi(I_t^i,y) \;\;\forall i\in [n]$ ~~~$\triangleright$ per-crop diffusion updates \newline
     $J_{t-1}\gets\texttt{MultiDiffuser}(\set{I^i_{t-1}}_{i=1}^n)$ ~~~~~~~~~ $\triangleright$ \eqref{e:closedform}
    }
    $\texttt{Panorama} \gets \mathcal{D}(J_0) $ ~~~$\triangleright$ Decode the panorama to RGB space 
    \newline
    \Output{ \texttt{Panorama}}
\end{algorithm}

Note that we can compute the per-crop diffusion updates in parallel (i.e., in a batch), resulting in total of ${\frac{{T \cdot n}}{b}}$ calls to the reference diffusion $\Phi$, where $b$ denotes the batch size.

\subsection{Bootstrapping (Sec.~\ref{sec:bootstrapping})}
\label{appendix:bootstrapping}
In case the user desires to maintain high fidelity to tight masks (see Fig. 4), we introduce a bootstrapping phase to our maps $F_i$ (see Eq. 9). Specifically, we pre-compute each $S_t$ as follows: we randomize an image $I\in [0,1]^{512x512x3}$ with a random constant RGB value, and encode it to Stable Diffusion latent space $S=\mathcal{E}(I)$, where $\mathcal{E}$ is the pre-trained encoder provided by the Stable Diffusion framework. Finally, we obtain $S_t$ by noising $S$ to the noise level of time-step $t$. That is, $S_t\sim \mathcal{N}$ where ${(\mu_t \cdot S,\sigma_t^2)}$, $\mu_t$ and $\sigma_t$ are the diffusion noise schedulers \cite{ddpm}. 

\end{document}